\pdfoutput=1

\documentclass[11pt]{article}

\usepackage[preprint]{acl}

\usepackage{times}
\usepackage{latexsym}
\usepackage{hyperref}
\usepackage{url}
\usepackage{booktabs}
\usepackage{comment}

\usepackage[skip=5pt]{caption}
\usepackage{enumitem}
\setlist[itemize]{noitemsep, topsep=1pt}

\usepackage{kotex}
\usepackage{amsmath,amsthm,amsfonts,amssymb,bm,stmaryrd}
\usepackage{algorithm}
\usepackage{algorithmicx}
\usepackage{algpseudocode}

\usepackage[T1]{fontenc}

\usepackage[utf8]{inputenc}

\usepackage{microtype}

\usepackage{inconsolata}

\usepackage{graphicx}

\title{\textit{CacheFocus}: Dynamic Cache Re-Positioning for Efficient Retrieval-Augmented Generation}

\renewcommand{\thefootnote}{\fnsymbol{footnote}}
\author{Kun-Hui Lee\footnote[1]{} \\
  Jeonbuk National University \\
  \texttt{dkghszkfhs@jbnu.ac.kr} \\\And
  Eunhwan Park\footnote[1]{} \\
  Jeonbuk National University \\
  \texttt{judepark@jbnu.ac.kr} \\\And
  Donghoon Han \\
  Seoul National University \\
  \texttt{dhk1349@snu.ac.kr} \\\AND
  Seung-Hoon Na\footnote[2]{} \\
  Jeonbuk National University \\
  \texttt{nash@jbnu.ac.kr} \\
}

\begin{document}
\maketitle
\footnotetext{\footnote[1]{} Equal contribution.}
\footnotetext{\footnote[2]{} Corresponding author.}
\renewcommand{\thefootnote}{\arabic{footnote}}

\begin{abstract}
Large Language Models (LLMs) excel across a variety of language tasks yet are constrained by limited input lengths and high computational costs. Existing approaches\textemdash such as relative positional encodings (e.g., RoPE, ALiBi) and sliding window mechanisms\textemdash partially alleviate these issues but often require additional training or suffer from performance degradation with longer inputs. In this paper, we introduce \textbf{\textit{CacheFocus}}, a method that enhances length normalization and reduces inference latency without any further training. Our approach leverages query-independent, offline caching to efficiently reuse a Context KV Cache Store. We address the amplification of abnormal token distributions problem by re-positioning cached keys and introducing Layer-Adaptive Cache Pruning to discard low-relevance caches during pre-filling. Additionally, our Adaptive Positional Allocation Strategy dynamically reassigns cache positions to maximize the use of the available positional encoding range. Experiments on the Natural Questions and TriviaQA datasets demonstrate that CacheFocus outperforms alternative methods even when inputs exceed the $4$K limit of the \texttt{LLaMA-2} model, emphasizing its practical effectiveness for long-context LLMs. Moreover, even with large maximum input length of \texttt{Qwen2}, the performance of CacheFocus shows that it maintains consistent performance even as the number of documents increases, effectively managing long-text generation without degradation.
\end{abstract}
\newcommand{\llamasevenb}{\texttt{LLaMA-2-7B-Chat}}
\newcommand{\qwenoneb}{\texttt{Qwen2-1.5B-Instruct}}
\newcommand{\qwensevenb}{\texttt{Qwen2-7B-Instruct}}

\section{Introduction}
\begin{figure}[!hbt]
    \centering
    \includegraphics[width=0.8\linewidth]{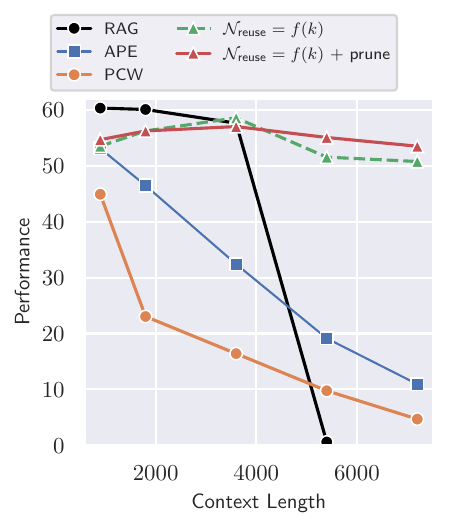}
    \caption{The performance of CacheFocus on NQ for \llamasevenb. This indicates that CacheFocus is not only adequate for extending input length but also showing robust performance.}
    \label{fig:proposal_method_best}
\end{figure}
Recent advancements in Large Language Models (LLMs) have demonstrated significant improvements in a wide range of language tasks, largely attributed to the effective encapsulation of knowledge within their extensive parameters during the pre-training. The reason beyond this success is that increasing the number of parameters allows us to improve generalization and performance in LLMs~\cite{hoffmann2022trainingcomputeoptimallargelanguage}. Additionally, Retrieval-Augmented Generation~(RAG) explicitly incorporates knowledge from retrieved documents, enhancing performance while mitigating the inherent hallucination of LLMs. However, LLMs continue to face significant challenges. The restricted input length limits the full utilization of knowledge of retrieved documents, with the performance and generalization capabilities of LLMs diminishing as input lengths increase. Moreover, the high computational costs required for LLMs pose substantial obstacles to their deployment in real-world applications.

Previous works for mitigating limitations, relative positional encodings such as RoPE~\cite{su2023roformerenhancedtransformerrotary} and ALiBi~\cite{Press2021trainshorttestlong} which are based on inter-token attention have been widely adopted to replace absolute positional encodings. Nonetheless, it has been observed that while generating text exceeding the input length, they could lead to model collapse and performance degradation, thereby necessitating additional training to optimize length generalization \cite{Ding2024longropeExtendingllmcontextwindow, Li2024extendingContextWindow, Chen2023clexContinuouslengthextrapolation, Zhu2023poseEfficientcontextwindow, hu2024longrecipeRecipeforEfficient}.
Moreover, applying a sliding window attention on the Transformer architecture could lead to effective memory usage~\cite{beltagy2020longformerlongdocumenttransformer}. Following from previous works, Parallel Context Windows~(PCW)~\cite{Ratner2023ParallelContextWindows} splits in-context examples or multiple documents into parallel windows, then merges the outputs while reusing a shared positional encoding range.
Although PCW is effective with up to $3$ windows, its performance degrades beyond that point, partly due to the duplication of tokens.
Some methods address this by keeping the parallelism unchanged but scaling attention weights, for example, through shared prefixes or heuristic factors~\cite{Hao2022structuredpromptingScalingInContext, yang2025apeFasterandLongerContextAugmented}, whereas others reduce the parallel load by pruning documents based on query-document relevance~\cite{Zhu2024AcceleratingInferenceofRetrievalAugmentedGeneration, sun2024blockattentionefficientrag, yen-etal-2024-long, li2024focusllmpreciseunderstandinglong}. Nonetheless, these solutions often rely on additional training or heuristic attention manipulations, limiting their ability to generalize efficiently in longer contexts.


To this end, we propose \textbf{\textit{CacheFocus}}, a simple yet effective method to enhance length normalization and inference latency without \textit{any} further training. First, we perform a query-independent parallel operation for \textit{offline} caching to facilitate the reuse of the \textit{Context KV Cache Store}. we directly mitigate both amplification of abnormal token distributions and parallel issues through \textit{Cache Re-Positioning}, i.e., shifting keys to different positions in the encoding space. This effectively scales the attention softmax distribution without introducing additional training or complex heuristics. Finally, we propose a \textit{Layer-Adaptive Cache Pruning}, where the attention scores between the query and each document are calculated at each layer, and caches for documents with low attention scores are removed while pre-filling. This may lead to discontinuities in the key's positional encodings. To remedy this, we introduce \textit{Adaptive Positional Allocation Strategy}, which adjusts the positions based on continuity or relevance.

We confirm the effectiveness of CacheFocus by extensive experiments on both the Natural Questions (NQ) and TriviaQA datasets using \texttt{LLaMA-2}, consistently surpassing alternative approaches varying input length, including scenarios exceeding $4$K tokens. Additionally, even with \texttt{Qwen2}’s expanded input length, our findings indicate that CacheFocus maintains stable performance as more documents are increased, effectively handling long-text generation without any remarkable performance drop in quality.


\section{CacheFocus}
\begin{figure*}[!hbt]
    \centering
    \includegraphics[width=0.9\linewidth]{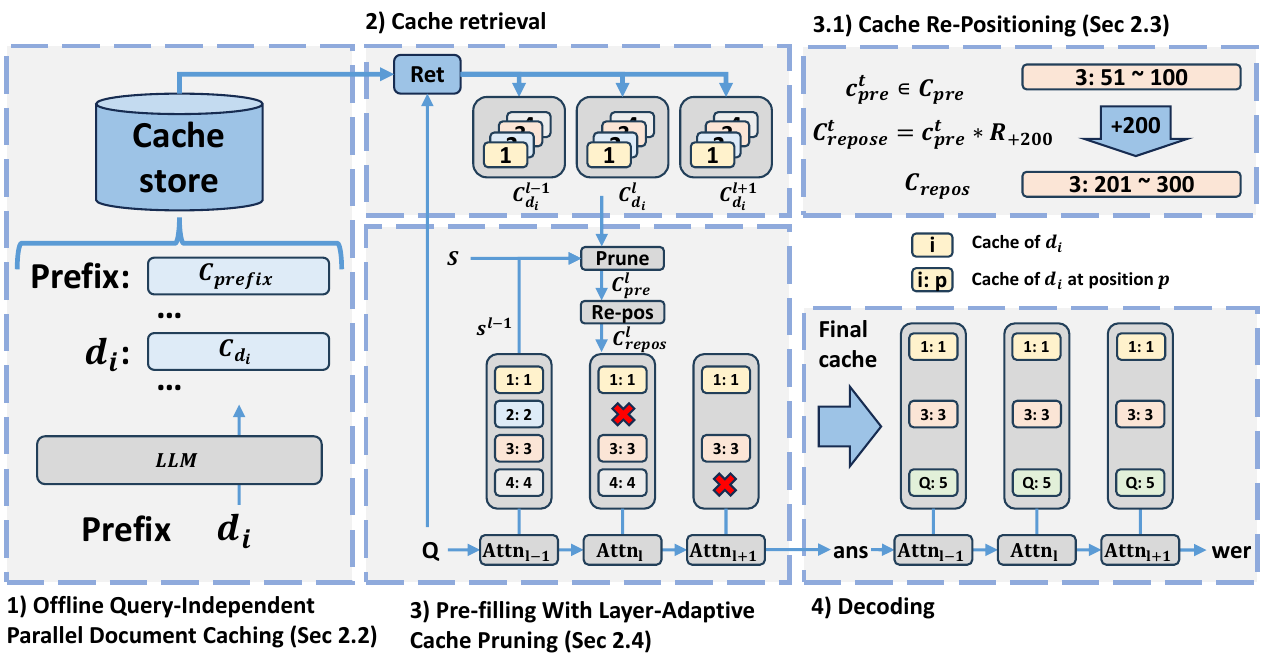}
    \caption{
    An overall architecture of \textbf{\textit{CacheFocus}}: 1) Offline Query-Independent Parallel Document Caching~(\S\ref{sec:query_indepedent_parallel_document_caching}): Documents are splitted into fixed-length passages and cached along with a shared prefix; 2) Cache Retrieval: Given a query, relevant passages and their caches are retrieved by a retriever; 3) Pre-filling with Layer-Adaptive Cache Pruning~(\S\ref{sec:layer_adaptive_cache_pruning}): The pre-computed caches are positioned within the model's positional encoding range~(\S\ref{sec:cache_reposition}), and \textit{Layer-Adaptive Cache Pruning}~(\S\ref{sec:layer_adaptive_cache_pruning}) is applied at specific layers based on accumulated attention scores, allowing the model to select semantically relevant documents; 4) Decoding: After pre-filling, the final caches are re-positioned according to the pruned caches via \textit{Adaptive Positional Allocation Strategy}~(\S~\ref{sec:adaptive_positional_allocation_strategy}), and the model proceeds with decoding, thereby obtaining reduced computational cost and high-quality context.}
    
    \label{fig:process}
\end{figure*}

\begin{figure}[!t]
    \centering
    \small
    \begin{algorithm}[H]
        \caption{Pre-filling \& Decoding}
        \begin{algorithmic}[1]
            \State \textbf{Input:} $X = \{x_1, x_2, ..., x_n\}$
            \State \textbf{Output:} $Y = \{y_1, y_2, ..., y_m\}$
            \State \textbf{Phase 1: Pre-filling}
            \State $\mathcal{C} \gets \emptyset$ (\textit{Empty}) 
            \State $\text{Output}, \mathcal{C} \gets \text{LLM}(\mathcal{C}, X)$ 
            \State $y_1 \gets \text{argmax}(\text{Output}[-1])$ 
            \State \textbf{Phase 2: Decoding}
            \For{$t = 2$ to $m$}
                \State $\text{Output}, \mathcal{C} \gets \text{LLM}(\mathcal{C}, y_{t-1})$ 
                \State $y_t \gets \text{Argmax}(\text{Output})$ 
            \EndFor
            \State \Return $Y$ 
        \end{algorithmic}
    \end{algorithm}
    \hfill
    \vspace{-2em}
    \addtocounter{algorithm}{-1}
    \captionof{algorithm}{An illustration of pre-filling and decoding phases in auto-regressive models. The pre-filling phase processes the input sequence to initialize the key-value cache, while the decoding phase iteratively generates tokens using the cached context.}
    \label{alg:background_prefill_decode}
\end{figure}

\subsection{Background}
In this section, we present the essential components of CacheFocus, focusing on three key areas: the scaled dot-product attention~\cite{Vaswani2017selfattention}, the pre-filling and decoding, and the use of Rotary Position Embedding~(RoPE)~\cite{su2023roformerenhancedtransformerrotary}.

\paragraph{Scaled Dot-Product Attention.} The attention mechanism measures the relevance between a query and a set of key states, then computes a weighted sum of the corresponding value states. Suppose that we have key-value states $\{\mathsf{K}, \mathsf{V}\}$, formed by concatenating the cached representation $\{\mathsf{K}_{\text{past}}, \mathsf{V}_{\text{past}}\}$ from previous steps with those computed from the current input $\{\mathsf{K}_{\text{current}}, \mathsf{V}_{\text{current}}\}$. Given a query $\mathsf{Q}$, the scaled dot-product attention is then calculated as follows:
\begin{equation}
     \textsf{Attn(Q, K, V)} = \textsf{softmax}\Big(\frac {\mathsf{Q} \cdot \mathsf{K}^T}{\sqrt{d}}\Big) \cdot \mathsf{V},
\end{equation}
where $d$ is the dimensionality of the hidden states\footnote{Note that casual masking is generally used to prevent attention to future token, we omit these details here for brevity.}.

\paragraph{Pre-filling \& Decoding.} Auto-regressive models generally split text generation into two phases: pre-filling and decoding, as shown in Algorithm~\ref{alg:background_prefill_decode}. During pre-filling, the input sequence $\mathcal{X}$ initializes the key-value cache $\mathcal{C}$ and produces the first output token. In the decoding phase, each newly generated token is fed back into the model along with the updated cache until a termination condition is met, which produces the final output $\mathcal{Y}$.

\paragraph{Rotary Position Embedding~(RoPE).} Recent LLMs replace absolute positional encodings with RoPE which encodes relative positions by rotating each two-dimensional slice of the query and key vectors with position-dependent angles. Specifically, let $r_i$ be the rotation angle for position $i$, and the define rotation matrix as follows:
\begin{equation}
R_i =
\begin{bmatrix}
    \cos (r_i) & -\sin (r_i) \\
    \sin (r_i) & \cos (r_i)
\end{bmatrix}.
\label{eq:rope_rotation_matrix}
\end{equation}
Given unrotated key and query vectors $k_0, q_0\in \mathbb{R}^2$, their rotated forms become $k_i=R_ik_0$ and $q_j=R_jq_0$. Note that the dot product $q_j^T k_i$ reflects the relative position $(j-i)$, allowing RoPE to capture positional relationships more flexibly than absolute positional encodings, particularly for the context of variable length.

\subsection{Query-Independent Parallel Document Caching}
\label{sec:query_indepedent_parallel_document_caching}
In auto-regressive models, subsequent tokens to be generated do not affect prior tokens. Based on this, we employ a query-independent parallel document caching by placing the query after the documents, meaning that document caches could be pre-computed in \textit{offline}. We first compute a shared prefix cache $\mathcal{C}_{\text{prefix}}$ using a shared prefix $\text{p}$ to mitigate the duplicated attention sink problem as follows:
\begin{equation}
    \mathcal{C}_{\text{prefix}} \leftarrow \mathsf{LLM}(\text{p})
\end{equation}
For each document $d_i$, we derive its cache by combining $\mathcal{C}_{\text{prefix}}$ with $d_i$:
\begin{equation}
    \mathcal{C}_{d_i} \leftarrow \mathsf{LLM}(\mathcal{C}_{\text{prefix}}, d_i)
\end{equation}
Note that the cache for each document is pre-comupted and stored Context KV Cache Store\footnote{Here, in the case of caching, $\mathsf{LLM}(\cdot)$ is a function that could take a past cache and a token sequence as input, and produce the corresponding cache for that sequence.}.

\subsection{Cache Re-Positioning}
\label{sec:cache_reposition}
As it is theorized that two-dimensional rotation matrices have inverses that are simply their transposes, RoPE utilizes this property to revert a key vector to its \textit{unrotated form} and subsequently re-apply RoPE for a new target position. Formally, suppose that $k_0 \in \mathbb{R}^{2}$ is the unrotated key vector and ${k_i=R_ik_0}$ its encoding at position $i$. Thus, the re-positioning follows as:
\begin{equation}
    k_j = R_j R_i^T k_i, \quad k_j = R_j k_0
    \label{eq:cache_reposition}
\end{equation}
Thanks to this property, recomputing RoPE enables flexible re-positioning of cached keys freely, effectively reusing the model's available positional encoding range and mitigating amplification of attention. Note that we leverage this mechanism to re-position cached keys in document caches before computing query -- document attention, thereby ensuring optimal positional encoding alignment.

\subsubsection{Adaptive Re-Positioning}
Previous works~\cite{Ratner2023ParallelContextWindows,Hao2022structuredpromptingScalingInContext, zhu2025acceleratingInferenceof} reuse positional IDs across windows by either attending to multiple documents within a single window or assigning one window per document, which often leads to inefficient utilization of the available positional encoding space, while leaving many encodings unused for short documents and hindering effective caching when documents share a window. Different from previous works, we dynamically adjust the positional encodings of cached keys to maximize the usage of the encoding space while mitigating the attention amplification associated with excessive reuse of positional IDs, in the term of \textit{Adaptive Re-Positioning}.

Given the model's available positional encoding length $L$ and the cache length $l_\mathcal{C}$, the maximum number of caches that can be assigned unique positions without reuse is 
$\frac{L}{l_\mathcal{C}}$.
For a retrieved set of $k$ caches, the required number of reuses for positional encodings is calculated as:
\begin{equation}
\mathcal{N}_{\text{reuse}} = \operatorname{ceil}\left(\frac{k}{L/l_\mathcal{C}}\right).
\label{eq:n_reuse_fk}
\end{equation}
We split the $k$ caches into $\mathcal{N}_{\text{reuse}}$ groups and assign sequential positions to the caches within each group via Cache Re-Positioning. These repositioned caches are then used for attention computation, ensuring optimal utilization of the positional encoding space while minimizing the adverse effects of excessive reuse.

\subsection{Layer-Adaptive Cache Pruning}
\label{sec:layer_adaptive_cache_pruning}
\begin{figure}[!t]
    \centering
    \small
    \begin{algorithm}[H]
        \caption{Layer-Adaptive Cache Pruning}
        \begin{algorithmic}[1]
            \State \textbf{Input:} $\mathcal{C}_{\text{prefix}}^l$,
            \State $\mathcal{C}_{d_i}^l (d_i \in D, i \in [1, k], l \in [1, L])$
            \State {\textbf{Init:}}
            \State $ids \gets \{i| d_i \in D\}$
            \State $ids_{sort} \gets \text{argsort}_i([\text{score}_{\text{ret}}(\mathcal{Q}, d_i)| i \in ids])$
            \State $\mathcal{S} \gets [s_i| s_i=0, i \in ids]$
            \State $\mathcal{H}^0 \gets \mathcal{Q}$
            \State \textbf{Phase 1: Pre-filling}
            \For{$l = 1$ to $L$}
                \State $\mathcal{C}_{\text{pre}}^l \gets \text{concat}([\mathcal{C}_{\text{prefix}}^l]+[\mathcal{C}_{d_i}^l| i \in ids])$
                \State $\mathcal{C}_{\text{repos}}^l \gets \text{reposition}(\mathcal{C}_{\text{pre}}^l, ids_{sort})$
                \State $\mathcal{C}_{\mathcal{Q}}^l, \mathcal{H}^l, s^l \gets \text{layer}_l(\mathcal{C}_{\text{repos}}^l, \mathcal{H}^{l-1})$
                \State $\mathcal{S} \gets \mathcal{S} + s^l$
                \If{$l \mod n=0$}
                    \State $ids_{sort} \gets \text{argsort}_i([s_i|s_i \in \mathcal{S}])$
                    \State $ids \gets \text{prun}(ids_{sort})$
                \EndIf
            \EndFor
            \State \textbf{Phase 2: Final Re-Positioning}
            \For{$l = 1$ to $L$}
                \State $\mathcal{C}_{\text{pre}}^l \gets \text{concat}([\mathcal{C}_{\text{prefix}}^l]+[\mathcal{C}_{d_i}^l| i \in ids])$
                \State $\mathcal{C}_{\text{repos}}^l \gets \text{reposition}(\mathcal{C}_{\text{pre}}^l, ids_{sort})$
                \State $\mathcal{C}_{\text{final}}^l \gets \text{concat}(\mathcal{C}_{\text{repos}}^l, \mathcal{C}_\mathcal{Q}^l)$
            \EndFor
            \State \textbf{Phase 3: decoding}
            \State $Y \gets \text{LLM}(\mathcal{C}_{\text{final}}^{all})$
            \State \Return $Y$
        \end{algorithmic}
    \end{algorithm}
    \hfill
    \vspace{-2em}
    \addtocounter{algorithm}{-1}
    \captionof{algorithm}{An illustration of Layer-Adaptive Cache Pruning, which progressively aggregates attention scores across layers to assess the semantic relevance of cached document and prune for attending document with higher semantic relevance.}
    \label{alg:method:LayerAdaptiveCachePruning}
\end{figure}
Intuitively, directing attention toward caches with higher semantic relevance would enhance long-text generation, while disregarding those with lower relevance. To this end, we introduce a layer-adaptive pruning technique that removes caches deemed less relevant based on their attention scores\footnote{At each layer, attention score is progressively computed and accumulated over the document-wise in the attention map, yielding an aggregated document-wise attention score.}. Given a query $\mathcal{Q}$, we retrieve a set of relevant documents $\mathcal{D} = \{d_i \mid 1 \leq i \leq k\}$ along with their respective caches $\mathcal{C}_{d_i}$ and a shared prefix cache $\mathcal{C}_{\text{prefix}}$. We then perform a pre-filling step, during which, at every $n\text{-th}$ layer, a pruning operation is executed.

Suppose that $\mathsf{reposition}(\cdot)$ as introduced in \S\ref{sec:cache_reposition} is a function that adjusts the positional encodings of precomputed caches, outputting a re-positioned cache $\mathcal{C}_{\text{repos}}$.
For each layer $l$, we construct a cache set $\mathcal{C}_{\text{pre}}^l$ based on the current cache IDs, concatenate these caches, and apply $\mathsf{reposition}(\cdot)$ to obtain $\mathcal{C}_{\text{repos}}^l$.
We then feed $\mathcal{C}_{\text{repos}}^l$ together with the hidden state $\mathcal{H}^{l-1}$ into layer $l$, which produces a new hidden state $\mathcal{H}^l$ and a relevance score $\mathcal{S}$ for each cache. At every $n\text{-th}$ layer, caches associated with low scores in $\mathcal{S}$ are pruned from the current cache set.

After processing through all layers, a final re-positioning step is performed using the remaining cache IDs. This step prunes caches from earlier layers accordingly and applies $\mathsf{reposition}(\cdot)$ to yield a combined final cache $\mathcal{C}_{\text{final}}^{\text{all}}$. The final cache is then used for decoding, ensuring that only the most relevant caches contribute to the generation process.

\subsection{Adaptive Positional Allocation Strategy}
\label{sec:adaptive_positional_allocation_strategy}
\begin{figure}[!hbt]
    \centering
    \includegraphics[width=0.8\linewidth]{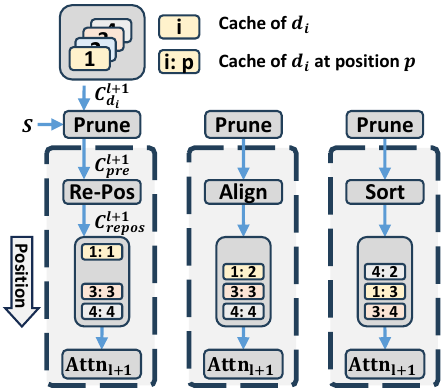}
    \caption{An illustration of ``align'' and ``sort'' strategies in \S\ref{sec:adaptive_positional_allocation_strategy}. The ``align'' strategy simply assigns cache positions to available spots in the positional encoding space, whereas the ``sort'' strategy allocates these positions based on attention scores. Note that all strategies manipulates cache positions closed to query inspired by \cite{liu2024LostintheMiddle}.}
    \label{fig:Cache_Pruning}
\end{figure}
\textit{Layer-Adaptive Cache Pruning}, as introduced in \S\ref{sec:layer_adaptive_cache_pruning}, results in unused positions within the available positional encoding range. Inspired by \cite{liu2024LostintheMiddle}, which demonstrates that closer proximity between a query and relevant documents enhances performance, we propose two strategies to reassign new positions to the remaining caches after pruning: the \textbf{\textit{Dynamic Positional Allocation Strategy}} and the \textbf{\textit{Attention-Guided Allocation Strategy}}\footnote{For brevity, the Dynamic Positional Allocation Strategy and the Attention-Guided Allocation Strategy are referred to as ``align'' and ``sort'', respectively.}. These strategies aim to maximize the utilization of the positional encoding range and improve overall performance. As shown in Figure~\ref{fig:Cache_Pruning}, the \textbf{\textit{Dynamic Positional Allocation Strategy}} assigns new positions dynamically to maximize encoding space usage, while the \textbf{\textit{Attention-Guided Allocation Strategy}} reorders caches based on their aggregated attention scores during pruning.

\section{Experiments}

\begin{figure*}[!htb]
    \centering
    \includegraphics[width=0.9\linewidth]{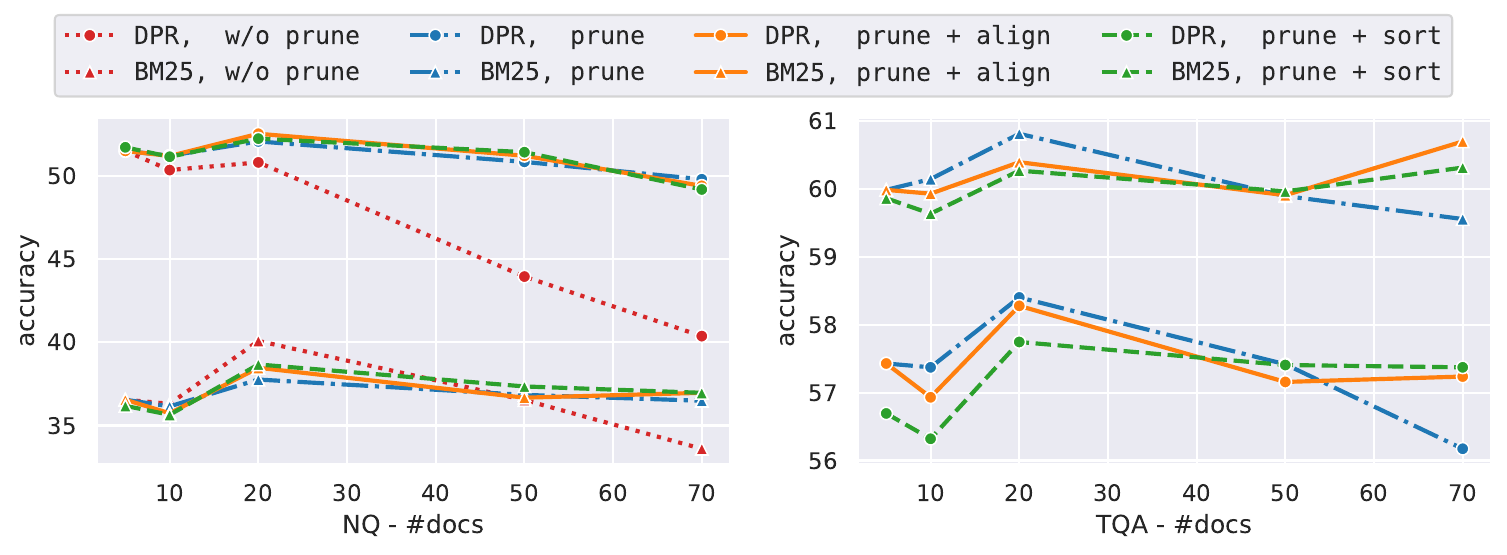}
    \caption{The performance of \textbf{\textit{CacheFocus}} on \qwensevenb~for NQ and TQA. Note that since the maximum input length of \qwensevenb~is 32K, we set \(\mathcal{N}_{\text{reuse}} = 1\), meaning that all windows are unique positions exactly once.}
    \label{fig:nq_tqa_qwen2_7B_prune}
\end{figure*}

\subsection{Settings}

The futher detail of dataset, retrieval, and setup of CacheFocus could be found in Appendix.
\paragraph{Large Language Models.} We employ the \texttt{Qwen2-\{1.5, 7\}B-Instruct}~\cite{Yang2024qwen2TechnicalReport} and \texttt{LLaMA-2-7B-Chat}~\cite{Touvron2023llama2Openfoundation} for LLMs.

\paragraph{Evaluation Metric.} Following from \cite{liu2024LostintheMiddle}, we evaluate the performance using accuracy whether at least one correct answer appears in the model's output.

\paragraph{Baseline.} Our baseline is based on the PCW approach \cite{Ratner2023ParallelContextWindows}, with the modification of using the system prompt as the shared prefix instead of \texttt{[BOS]}. We employ zero-shot settings, with one document per window and greedy decoding. Moreover we also compared our method with RAG, and APE~\cite{yang2025apeFasterandLongerContextAugmented} in the case of \llamasevenb. 

\subsection{Main Results}

Figure~\ref{fig:proposal_method_best}, \ref{fig:nq_tqa_qwen2_7B_prune} present the experimental results on the NQ and TQA datasets using \qwensevenb, as well as on NQ using \llamasevenb, respectively.

\paragraph{\textit{CacheFocus} remains effective even beyond the model's maximum input length.} Figure~\ref{fig:proposal_method_best} illustrates that RAG based on \llamasevenb~experiences a substantial performance drop when the maximum input length (i.e. $4$K tokens) is exceeded. Although APE and PCW partially mitigate this decline, they still exhibit gradual performance degradation as input length increases. In contrast, CacheFocus consistently maintains robust performance even when the input surpasses the maximum length, thereby demonstrating its efficacy in handling extended inputs.

\paragraph{\textit{Layer-Adaptive Cache Pruning} presents a large room for Extended Contexts.} Even with large maximum input length~(i.e. $32$K tokens) of \qwensevenb, Figure~\ref{fig:nq_tqa_qwen2_7B_prune} indicates that performance on NQ declines once more than $20$ documents are provided without pruning. This suggests that semantically irrelevant caches might act as noise during inference. As shown in Figure~\ref{fig:nq_tqa_qwen2_7B_prune}, pruning discards low relevance caches, thereby preserving higher-quality context, meaning that it would lead to stable performance. Moreover, as the number of documents increases, the gap in accuracy between ``w/o prune'' and ``prune'' widens, meaning that pruning shows large rooms for further handling extensive parallel context. These results also confirm that managing the relevance of caches is essential for preserving stable performance, rather than simply relying on a large input capacity to handle all retrieved documents.

\subsection{Analysis}
\begin{table*}[ht]
    \centering
    \scalebox{0.7}{
    \begin{tabular}{l c | cccc | cccc}
    \toprule
         & & \multicolumn{4}{c|}{\textbf{DPR-based}} & \multicolumn{4}{c}{\textbf{BM25-based}} \\
    \textbf{Method} & \textbf{Layer} & \textbf{R@5} & \textbf{R@10} & \textbf{R@20} & \textbf{R@50} & \textbf{R@5} & \textbf{R@10} & \textbf{R@20} & \textbf{R@50} \\
    \midrule
    \midrule
    \textbf{(Baselines)} & & & & & & & & & \\
    DPR & -- & \textbf{0.7075} & 0.7925 & 0.8575 & \textbf{0.9125} & -- & -- & -- & -- \\
    BM25 & -- & -- & -- & -- & -- & 0.2500 & 0.3488 & 0.4387 & 0.5337 \\
    \midrule
    \textbf{(Pruning Methods)} & & & & & & & & & \\
    Pruning & 4  & 0.6987 & 0.7875 & 0.8488 & 0.9050 & 0.2537 & 0.3425 & 0.4200 & 0.5350 \\
    Pruning & 16 & 0.6937 & 0.7913 & \textbf{0.8600} & \textbf{0.9125} & 0.2700 & 0.3650 & 0.4688 & \textbf{0.6225} \\
    Pruning & 28 & 0.6925 & 0.7850 & 0.8550 & \textbf{0.9125} & 0.2712 & 0.3713 & 0.4850 & \textbf{0.6225} \\
    \midrule
    Pruning + Dynamic Positional Allocation & 4  & 0.6987 & 0.7875 & 0.8488 & 0.9050 & 0.2537 & 0.3425 & 0.4200 & 0.5350 \\
    Pruning + Dynamic Positional Allocation & 16 & 0.6937 & \textbf{0.7925} & 0.8575 & \textbf{0.9187} & 0.2700 & 0.3675 & 0.4713 & 0.5925 \\
    Pruning + Dynamic Positional Allocation & 28 & 0.6937 & 0.7887 & 0.8562 & \textbf{0.9187} & \textbf{0.2800} & 0.3762 & \textbf{0.4913} & 0.5925 \\
    \midrule
    Pruning + Attention-Guided Allocation & 4  & 0.6987 & 0.7875 & 0.8488 & 0.9050 & 0.2537 & 0.3425 & 0.4200 & 0.5350 \\
    Pruning + Attention-Guided Allocation & 16 & 0.6875 & 0.7863 & 0.8500 & \textbf{0.9137} & 0.2625 & 0.3600 & 0.4487 & 0.6025 \\
    Pruning + Attention-Guided Allocation & 28 & 0.6887 & 0.7837 & 0.8512 & \textbf{0.9137} & 0.2737 & \textbf{0.3862} & 0.4900 & 0.6025 \\
    \bottomrule
    \end{tabular}
    }
    \caption{The retrieval performances for DPR-based and BM25-based retrieval under different methods (\S\ref{sec:layer_adaptive_cache_pruning}, \S\ref{sec:adaptive_positional_allocation_strategy}) and layers ($4$, $16$, $28$) on \qwenoneb. Baseline rows (DPR, BM25) indicate retrieval performance without pruning.}
    \label{tab:unified_retrieval_results}
\end{table*}

\paragraph{The effect of \textbf{\textit{Adaptive Positional Allocation Strategy}}.} Table~\ref{tab:unified_retrieval_results} presents the re-ranking performance on NQ for different methods\textemdash pruning alone and pruning combined with the adaptive positional allocation strategy\textemdash using both DPR-based and BM25-based retrieval on \qwenoneb. For DPR-based retrieval, which is already trained on NQ, re-ranking yields only marginal improvements in R@50. In contrast, BM25-based retrieval, relying soley on lexical matching and neglecting semantic relevance, indicates a substantial improvement of $2.12$ in R@5 with pruning alone. Moreover, when the adaptive positional allocation strategy is applied, additional improvements of $3.0$ and $2.37$ points in R@5 are observed for the respective strategies compared to BM25-based, implying that our approach could significantly enhance performance, particularly when the retriever's semantic capabilities are limited.

\paragraph{Pruning with positional allocation strategy improves overall performance.} Since Layer-Adaptive Cache Pruning demonstrates effective in discarding low-relevance caches, additional improvements could be achieved by dynamically allocating the remaining caches. In particular, \textbf{\textit{Dynamic Positional Allocation}} and \textbf{\textit{Attention-Guided Allocation}} strategies provide additional benefits when the retrieval models is not fully aligned with the target dataset. For instance, when evaluating TQA, both strategies yield remarkable gains over ``prune'' by allocating new positions, compensating for potential mismatches in the retrieval's original ranking. Conversely, in NQ\textemdash where the retriever is trained directly\textemdash the ranking is already well trained and thus the advantage of strategies is marginal. This implies that while pruning alone helps filter out noisy caches, combining it with strategies could be helpful for datasets where the initial retrieval model is sub-optimal.

\paragraph{Is \textit{Adaptive Re-positioning} adequate for extending context?}
\begin{figure}[ht]
    \centering
    \includegraphics[width=0.8\linewidth]{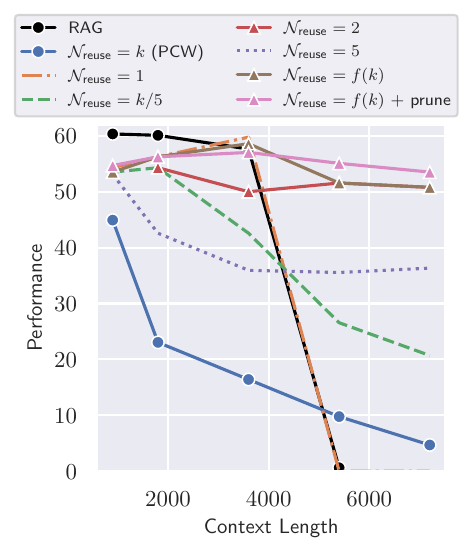}
    \caption{The preformance of CacheFocus without applying \S\ref{sec:layer_adaptive_cache_pruning} on NQ for \llamasevenb. It confirms that Cache Re-Positioning is essential component for extending maximum input length.}
    
    \label{fig:proposal_method}
\end{figure}
Figure~\ref{fig:proposal_method} indicates the performance comparison across varying the number of reused positions $\mathcal{N_{\text{reuse}}}$. For instance, if there are $k=20$ documents, the $\mathcal{N}_{\text{reuse}}=\frac{k}{5}$ implies that $5$ windows are assigned continuous positions (i.e. reused $4$ times). To investigate the effect of adaptive re-positioning, we explore varing $\mathcal{N}_{\text{reuse}}$ as follows:
\begin{itemize}
    \item \textbf{$\mathcal{N}_{\text{reuse}}=k$} (PCW~\cite{Ratner2023ParallelContextWindows}): All windows are placed in parallel, increasing reuse proportionally with the number of documents.
    
    \item \textbf{$\mathcal{N}_{\text{reuse}}=1$}: All windows use \textit{unique} positions exactly once. While this significantly boosts performance relative to $\mathcal{N}_{\text{reuse}}=k$ and similar to performance of na\"ive RAG, it could lead to decrease the performance if the positional encoding ranges is exceeded.

    \item \textbf{$\mathcal{N}_{\text{reuse}}=k/5$}: Reusing positions every $5$ documents might lead to performance degradation as the reuse count grows, however remaining robust even when input exceed the maximum length.

    \item \textbf{$\mathcal{N}_{\text{reuse}}=5$}: Fixing the number of $\mathcal{N}_{\text{reuse}}$ could achieve stable performance as $k$ increases, and using only $2$ reuse times could yield even better results.
\end{itemize}

As indicated in Figure~\ref{fig:proposal_method}, $\mathcal{N}_{\text{reuse}}=f(k)$ (Eq~\ref{eq:n_reuse_fk}) adjusts the reuse count based on the number of documents, thereby successfully capturing the benefits of both parallel and sequential positioning.

\paragraph{Inference Latency.}
\begin{table}[ht]
    \centering
        \scalebox{0.8}{\begin{tabular}{ll|cc|c}
        \hline
        \textbf{Length (\#doc)} & \textbf{Model} & \textbf{Prefill} & \textbf{Decode} & \textbf{Total}\\
        \hline \hline
                & naive     & 0.174 & 3.268 & 3.442 \\
        2K~(10)  & w/o cache & 0.652 & 3.066 & 3.718 \\
                & w/ cache  & 0.073 & 3.053 & 3.126 \\
                & w/ prune  & \textbf{0.075} & \textbf{2.995} & \textbf{3.070} \\
\hline          & naive     & 0.454 & 4.157 & 4.611 \\
        4K~(20)  & w/o cache & 1.260 & 3.594 & 4.854 \\
                & w/ cache  & 0.107 & 3.566 & 3.673 \\
                & w/ prune  & \textbf{0.095} & \textbf{3.067} & \textbf{3.162} \\
\hline          & naive     & 1.953 & 6.490 & 8.443 \\
        8K~(40)  & w/o cache & 2.549 & 4.764 & 7.313 \\
                & w/ cache  & 0.199 & 4.761 & 4.960 \\
                & w/ prune  & \textbf{0.154} & \textbf{3.476} & \textbf{3.630} \\
        \hline
    \end{tabular}}
    \caption{A table of $100$-token generation time analysis for the \llamasevenb~under varying settings. This indicates prefill, decode and total time for na\"ive without and with cache, and with pruning at input lengths of $2$K, $4$K, and $8$K.}
    \label{tab:cache_time}
\end{table}
Table~\ref{tab:cache_time} presents 100-token generation times for the \llamasevenb~model under varying cache management settings. In our analysis, we define the following:
\begin{itemize}
    \item $L$: Total cache length.
    \item $l_c$: Length of each individual cache.
    \item $k$: Number of retrieved caches.
    \item $n$: Final number of caches after pruning.
    \item $q$ and $a$: Query and generation lengths, respectively.
\end{itemize}

The prefill time in the na\"ive setting increases significantly with input length. In contrast, when caching is employed, the prefill time remains nearly constant, reducing to $81$\% of the naive approach for a $2$K input and $53$\% for a $4$K input. Moreover, since token decoding time remains similar when operating on caches of comparable length, the overall inference time is significantly reduced (e.g., at a $4$K input, the total inference time is lowered to $81$\% of the na\"ive baseline).

These empirical results are supported by our complexity analysis. Cache loading operates in $\mathcal{O}(L)$. For the prefill phase, the complexities follows as:
\begin{equation}
\mathcal{O}(L^2),\quad \mathcal{O}(l_c^2 k + Lq),\quad \mathcal{O}(Lq),\quad \mathcal{O}(l_c n q) \nonumber
\end{equation}
For the decoding phase, the complexities follows as:
\begin{equation}
\mathcal{O}\Bigl(\frac{(2L+a)a}{2}\Bigr),\quad \mathcal{O}\Bigl(\frac{(2l_c n+a)a}{2}\Bigr) \nonumber
\end{equation}
These expressions demonstrate that our method significantly reduces computational overhead, especially as the input length increases. This implies that our method not only improves inference latency but also scales effectively with longer input sequences, making it promising for efficient long-text generation.

\section{Related Work}
Recent works in RAG have explored methods to efficiently process and incorporate large contexts. PCW~\cite{Ratner2023ParallelContextWindows} splits few-shot examples or documents into several windows processed in parallel, thereby reducing positional encoding overhead and removing cross-attention between windows. While performance tends to decrease when using more than $3$ windows, \textbf{\textit{CacheFocus}} demonstrates robust performance even with contexts longer than those covered by $3$ windows.

Various modifications to the attention mechanism have been proposed to address limitations in context relevance and distribution. For example, Structured Prompting~\cite{Hao2022structuredpromptingScalingInContext} scales attention values by $1/M$ (where $M$ is the number of windows), although it requires specific in-context learning examples. Similarly, APE~\cite{yang2025apeFasterandLongerContextAugmented} investigates additional scaling factors and temperature reduction, but it does not consider cache positioning based on the semantic relevance between query and document caches. SparseRAG~\cite{Zhu2024AcceleratingInferenceofRetrievalAugmentedGeneration} infers relevance scores to prune less pertinent documents, which requires an additional training process. XL3M~\cite{Wang2024XL3MaTrainingfreeFrameworkforLLMLength} and Superposition Prompting~\cite{Merth2024SuperpositionPromptingImprovingandAccelerating} adopt strategies that split and filter long inputs using probabilistic measures and Bayesian inference, respectively.

In contrast to these approaches~\cite{yang2025apeFasterandLongerContextAugmented, Zhu2024AcceleratingInferenceofRetrievalAugmentedGeneration, Wang2024XL3MaTrainingfreeFrameworkforLLMLength, Merth2024SuperpositionPromptingImprovingandAccelerating}, \textbf{\textit{CacheFocus}} not only repositions caches based on the semantic relevance between query and document caches but also aggregates attention scores in a layer-wise manner, thereby capturing multi-level contextual information and mitigating noise from individual layers, which could lead to improved stability and performance.

\section{Conclusion}
In this paper, we propose \textbf{\textit{CacheFocus}}, a framework designed to enhance long-text generation in LLMs by reducing inference latency and handling extended inputs without \textit{any} further training. We leverage Query-Independent Parallel Document Caching~(\S\ref{sec:query_indepedent_parallel_document_caching}) to pre-compute a Context KV Cache Store, and address the challenge of Attention Sink by introducing a Cache Re-Positioning mechanism~(\S\ref{sec:cache_reposition}) that dynamically adjusts positional encodings. Furthermore, Layer-Adaptive Cache Pruning~(\S\ref{sec:layer_adaptive_cache_pruning}) removes semantically irrelevant caches based on attention scores and Adaptive Positional Allocation Strategy~(\S\ref{sec:adaptive_positional_allocation_strategy}) consisting of Dynamical Positional Allocation and Attention-Guided Allocation further optimizes the assignment of positional encodings.

Experimental Results on the NQ and TQA datasets demonstrate that CacheFocus outperforms previous works~\cite{Ratner2023ParallelContextWindows, yang2025apeFasterandLongerContextAugmented}, even when input lengths exceed the maximum length of LLMs. Our analysis of time complexity confirms that our method significantly reduces computational overhead by lowering prefill and decoding times, while preserving performance in extended contexts. Finally, CacheFocus not only improves inference latency but also robustly scales with longer inputs, thereby paving the way for more efficient and accelerated long-text generation with LLMs.

\section*{Limitations}
While \textbf{\textit{CacheFocus}} shows promising results in handling long-text inputs with lower computational overhead, it also comes with several limitations as follows:
\begin{itemize}
    \item Offline Cache Pre-Computation: \textbf{\textit{CacheFocus}} relies on pre-computed caches for each document segment in an offline manner. This pre-computation could be memory-intensive and may not adapt well to rapidly changing or query-specific content. If the underlying documents are updated frequently, the offline caches could become stale, thereby necessitating frequent re-computation.
    \item Zero-Shot Experimental Setting: We evaluate \textbf{\textit{CacheFocus}} under zero-shot conditions. In contrast, PCW~\cite{Ratner2023ParallelContextWindows} was tested with few-shot examples on the NQ, which would offer additional performance gains. Future work could explore how few-shot or other prompting strategies influence the effective of \textbf{\textit{CacheFoucs}}.
    \item Segmentation of Short Passage: Currently, we segment document into relatively short passages for retrieval, which can simplify caching but might limit performance or applicability for very long, single-document inputs. Previous works have tacked extremely long documents within a single window, or included multiple passages per window. By contrast, our approach is less suitable when the content is not easily split or when each window dynamically combines multiple passages based on the query.
    \item Evaluation on a Limited Set of LLMs: Our experiments have been conducted using popular mdoels such as \texttt{LLaMA-2} and \texttt{Qwen2}. While these LLMs provide valuable performance, they might not fully represent the performance or capabilities of the latest state-of-the-art architectures. Future work should consider a broader range of LLMs to further validate the scalability and generalizability of \textbf{\textit{CacheFocus}}.
\end{itemize}

Overall, our limitations highlight potential room for future work, such as extending our method to more dynamic retrieval settings, incorporating different prompting strategies, experimenting with long and more varied input structures, and evaluating on architectures of state-of-the-art LLMs.

\bibliography{custom}

\appendix
\section{Instruction Format}
\begin{table}[ht]
    \centering
        \resizebox{\linewidth}{!}{\begin{tabular}{l|l}
        \hline
        shared prefix & <|im\_start|>system\textbackslash n\{system prompt\}<|im\_end|>\textbackslash n  \\
        \hline
        document   & \begin{tabular}[c]{@{}l@{}}<|im\_start|>retriever\textbackslash n Title\textbackslash n \{title\}\textbackslash n \\
         Passage\textbackslash n\{passage\}<|im\_end|>\textbackslash n\end{tabular} \\
        \hline
        question  & <|im\_start|>user\textbackslash n \{question\}<|im\_end|>\textbackslash n \\
        \hline
        gen\_prompt   & <|im\_start|>assistant\textbackslash n The answer is\\
        \hline
    \end{tabular}}
    \caption{The instruction format of \qwenoneb~and~\qwensevenb~}
    \label{tab:qwen_inst_form}
\end{table}
\begin{table}[ht]
    \centering
        \resizebox{\linewidth}{!}{\begin{tabular}{ll}
        \hline
        shared prefix & \begin{tabular}[c]{@{}l@{}}<s>[INST] \text{<}<SYS>\text{>}\textbackslash n\{system prompt\}\textbackslash n\text{<}</SYS>\text{>}\textbackslash n\textbackslash n \\
        \text{<}<CTX>\text{>}\textbackslash n \end{tabular} \\
        \hline
        document   & \text{<}<P>\text{>} Passage: (Title: \{title\}) \{passage\} \text{<}</P>\text{>}\textbackslash n \\
        \hline
        question  & \text{<}</CTX>\text{>}\textbackslash n\textbackslash nQuestion: \{question\} [/INST] \\
        \hline
        gen\_prompt   & Answer:\\
        \hline
    \end{tabular}}
    \caption{The instruction format of~\llamasevenb}
    \label{tab:llama_inst_form}
\end{table}
We employ the \texttt{Qwen2-1.5B-Instruct}, \texttt{Qwen2-7B Instruct}~\cite{Yang2024qwen2TechnicalReport}, and \llamasevenb~\cite{Touvron2023llama2Openfoundation} models. The instruction formats we used are described in Tables~\ref{tab:qwen_inst_form} and \ref{tab:llama_inst_form}.

\section{Dataset}
\begin{table}[ht]
    \centering
        \resizebox{\linewidth}{!}{\begin{tabular}{lcc}
        \hline
        \textbf{Task} & \textbf{\#train} & \textbf{\#valid}  \\
        \hline
        \hline
        Natural Question & 58,880 & 6,515 \\
        TriviaQA & 78785 & 8,837 \\
        \hline
        Wikipeida segmented per 100 words & \multicolumn{2}{c}{21,015,324} \\
        \hline
    \end{tabular}}
    \caption{The table of QA dataset statistics.}
    \label{tab:dataset_statistics}
\end{table}
We conduct an Open-Domain Question Answering task using the Natural Questions~(NQ)~\cite{Kwiatkowski2019Naturalquestionsabenchmark} and TriviaQA~(TQA)~\cite{Joshi2017TriviaQAaLargeScaleDistantlySupervisedChallenge} datasets. 

\section{CacheFocus Setup}
The hyper-parameters of Layer-Adaptive Cache Pruning~(\S\ref{sec:layer_adaptive_cache_pruning}) of CacheFocus in our experiments follows as:

\begin{itemize}
    \item The Number of Final Caches $k_{\text{finish}}$: This spcifies how many caches~(i.e., documents) remain afther the pre-filling phase. In our experiments, we set this value to $5$.
    \item $n$: Pruning is performed every $n\text{-th}$ layer. In our experiments, $n=4$.
\end{itemize}
Based on parameters, the number of caches pruned at each pruning layer is computed as:
\begin{equation}
    k_{\text{prune}} = \frac{k-k_{\text{finish}}}{L/n}, \nonumber
\end{equation}
where $k$ is the total number of retrieved caches and $L$ is the total number of layers of the LLMs. For instance, if $L=28,~n=4,~k=40,~k_{\text{finish}}=5$, we have $k_{\text{prune}}=\frac{40 - 5}{28 / 4} = 5$.

\section{Retrieval Models}
\begin{table}[!ht]
    \centering
    \makebox[\linewidth][c]{%
        \resizebox{1.0\linewidth}{!}{%
            \begin{tabular}{ll|ccccc|c}
                \hline
                \textbf{Task} & \textbf{Retriever} & \textbf{R@5} & \textbf{R@10} & \textbf{R@20} & \textbf{R@50} & \textbf{R@100} & \textbf{MRR@100} \\
                \hline 
                \hline
                NQ & BM25 & 0.268 & 0.350 & 0.427 & 0.528 & 0.602 & 0.180 \\
                   & DPR  & 0.615 & 0.710 & 0.790 & 0.865 & 0.900 & 0.442 \\
                \hline
                TQA & BM25 & 0.595 & 0.697 & 0.774 & 0.851 & 0.897 & 0.437 \\
                    & DPR  & 0.375 & 0.458 & 0.537 & 0.625 & 0.684 & 0.266 \\
                \hline
            \end{tabular}%
        }
    }
    \caption{The retrieval performance for DPR and BM25 on NQ and TQA datasets.}
    \label{tab:Retriever_mrr}
\end{table}

Using the Pyserini~\cite{Lin2021PyseriniaPythonToolkit} toolkit, we utilize data preprocessed according to the DPR~\cite{Karpukhin2020DensePassageRetrieval} method. We employ a passage collection segmented from Wikipedia in 100-word units and retrieve the top-k passages using both DPR and BM25. Table~\ref{tab:Retriever_mrr} indicates the performance of retrieval models on the validation dataset. We measured Recall at K~(R@K) and MRR@100 with Pyserini. Note that in the case of TQA, we excluded 2,077 examples from the validation set that lacked a relevant answer document.
\end{document}